\documentclass[11pt]{article}

\usepackage[preprint]{acl}

\usepackage{times}
\usepackage{latexsym}
\usepackage[T1]{fontenc}
\usepackage[utf8]{inputenc}
\usepackage{microtype}
\usepackage{inconsolata}
\usepackage{graphicx}
\usepackage{booktabs}
\usepackage{multirow}
\usepackage{amsmath}
\usepackage{url}
\usepackage[normalem]{ulem}
\usepackage{array}
\usepackage{subcaption}
\usepackage{enumitem}
\graphicspath{{figures/}}

\title{AskChem: Claim-Centered Infrastructure\\
for Chemistry Literature Synthesis}

\author{Bing Yan \\
  New York University \\
  \texttt{bing.yan@nyu.edu} \And
  Gregory Wolfe \\
  Matterstack, Inc. \\
  \texttt{gwolfe@matterstack.ai} \AND
  Stefano Martiniani \\
  New York University \\
  \texttt{stefano.martiniani@nyu.edu} \And
  Kyunghyun Cho \\
  New York University \\
  \texttt{kyunghyun.cho@nyu.edu}}

\begin{document}
\maketitle

\begin{abstract}
Chemistry literature synthesis often requires assembling specific findings scattered across many publications, yet existing literature-search systems primarily return ranked document lists.
As a result, scientists and AI agents need to locate relevant information, verify their provenance, and assemble cross-paper answers manually.
We present \textbf{AskChem}, a claim-centered infrastructure for cross-paper chemistry search. AskChem changes the unit of retrieval from the paper to the provenance-carrying claim: each paper is converted into atomic, typed claims, each grounded by a source DOI and a verbatim quote or an explicit evidence locator. Over this shared claim store, AskChem exposes complementary structures for search and synthesis: a stabilized faceted taxonomy for hierarchical retrieval and browsing, an evidence graph linking claims through relations, and an exploratory living taxonomy that situates indexed papers under scientific principles.
AskChem currently indexes 2.4M claims from 147K papers and provides a web interface, as well as REST, SDK, and MCP access for AI agents.
On AskChem-Bench, grounding a GPT-5.5 reader in AskChem yields 100\% resolvable DOIs, compared with 88.3\% without retrieval, and the highest citation density among five tested systems. AskChem is live at \url{https://askchem.org}.
\end{abstract}

\section{Introduction}

\begin{figure}[t]
    \centering
    \includegraphics[width=\linewidth]{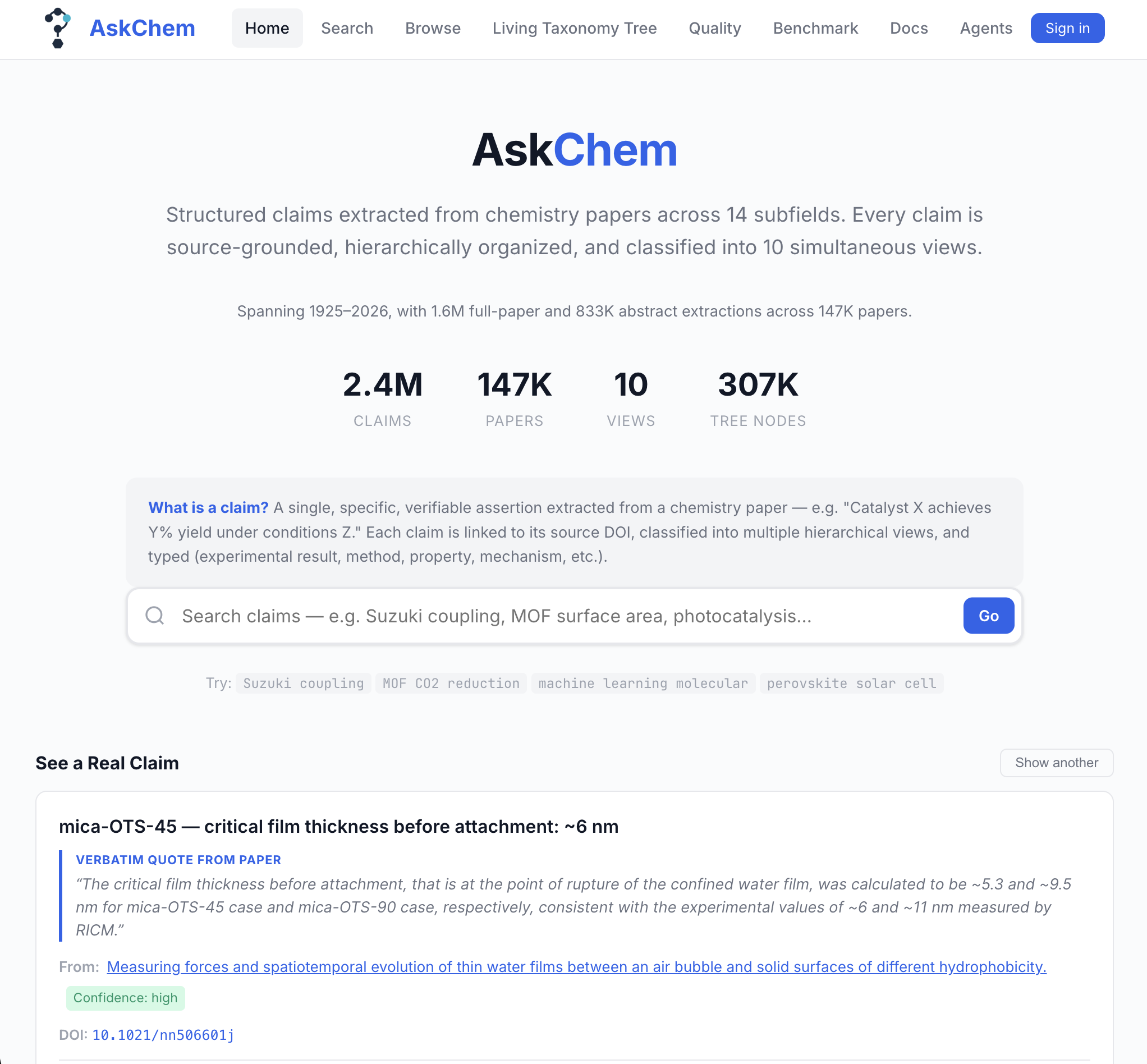}
    \caption{The \href{https://askchem.org}{AskChem} interface returns provenance-carrying claims.}
    \label{fig:website}
\end{figure}

\begin{figure}[t]
  \centering
  \includegraphics[width=\linewidth]{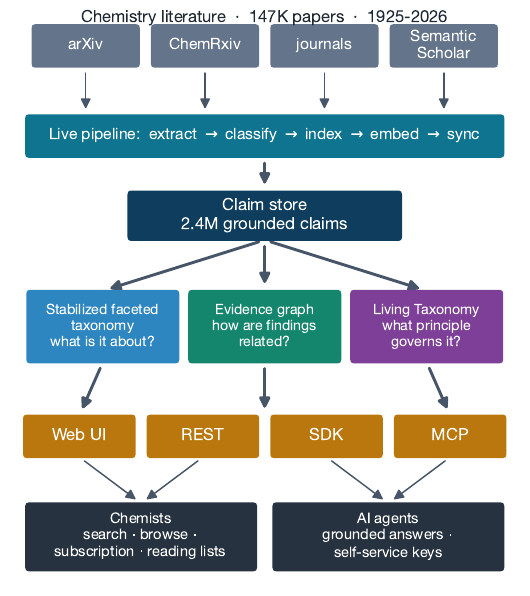}
  \caption{AskChem replaces papers with provenance-carrying claims as the retrieval unit, then exposes complementary structure over the shared claim store: a stabilized faceted taxonomy for search and browse, an evidence graph for cross-paper relations, and an exploratory living taxonomy organized by principles.}
  \label{fig:architecture}
\end{figure}

Chemists often ask questions whose answers must be assembled from specific findings across many papers, rather than retrieved from a single paper.
For example, when a chemist asks ``\emph{What electrocatalysts have been reported for CO$_2$ reduction to CO, and at what Faradaic efficiency?}'', the answer is distributed across many papers as individual claims about catalysts, reaction conditions, measurements, and mechanisms. Existing literature-search interfaces primarily return ranked lists of documents, leaving the scientist to open papers, locate relevant evidence, verify reported values, and assemble the answer by hand.

This document-centered interface is increasingly misaligned with how scientists and AI agents use the literature. LLM agents now survey papers and plan experiments \citep{boiko2023autonomous}, but inherit the limitations of their retrieval tools: document retrieval does not directly expose evidence at the claim level, the domain's organization, or the relationships between documents needed for synthesis. When asked to answer from parametric memory instead, an LLM may fabricate plausible-looking citations \citep{agrawal2023citations}.



\textbf{AskChem} addresses this mismatch by making the provenance-carrying scientific claim, rather than the paper, the central object for search, browsing, and agent access (Figure~\ref{fig:website}). A claim is an atomic, typed assertion extracted from a paper, grounded by a source DOI and a verbatim quote or an explicit evidence locator.
As the Segment Anything Model decomposed images into reusable regions \citep{kirillov2023sam}, we use LLMs to segment papers into claims.
By indexing claims directly, AskChem lets users retrieve specific findings, inspect their evidence, and assemble cross-paper answers without first reading and filtering whole documents.


AskChem then exposes three complementary structures over the shared claim store:

\paragraph{Faceted taxonomy: finding claims by what they concern.}
AskChem organizes claims along stabilized, corpus-induced facets such as reaction type, substance class, application, technique, mechanism topic, claim type, data, and time. These facets are not the scientific structure itself, but operational views that support hierarchical retrieval, browsing, grouping, and temporal exploration.

\paragraph{Evidence graph: connecting claims across papers.}
AskChem links claims through typed relations such as \texttt{supports}, \texttt{contradicts}, \texttt{extends}, and \texttt{derives\_from}. This graph helps users move from an individual finding to related evidence, follow how results are extended across papers, and surface potential conflicts.

\paragraph{Living Taxonomy: situating claims in scientific context.}
As an exploratory extension, AskChem places paper-grounded leaves beneath principles, theories, models, mechanisms, and phenomena. This principle-centered hierarchy provides an evolving overview of the indexed corpus, complementing the operational facets used for search.

\paragraph{Contributions.}
We present AskChem, a live claim-centered infrastructure for chemistry literature synthesis. 
Our contributions are fourfold:
\begin{enumerate}[leftmargin=*,nosep]
    \item a provenance-carrying claim representation deployed over 2.4M claims from 147K papers;
    \item complementary structures over the shared claim store, including a stabilized faceted taxonomy for retrieval and browsing, an evidence graph for claim-level navigation, and an exploratory principle-centered Living Taxonomy;
    \item human- and agent-facing interfaces through a web UI, REST API, SDK, and MCP server;
    \item AskChem-Bench, a cross-paper chemistry search evaluation suite for measuring citation groundedness and relevance.
\end{enumerate}

\begin{figure*}[t]
  \centering
  \includegraphics[width=0.7\textwidth]{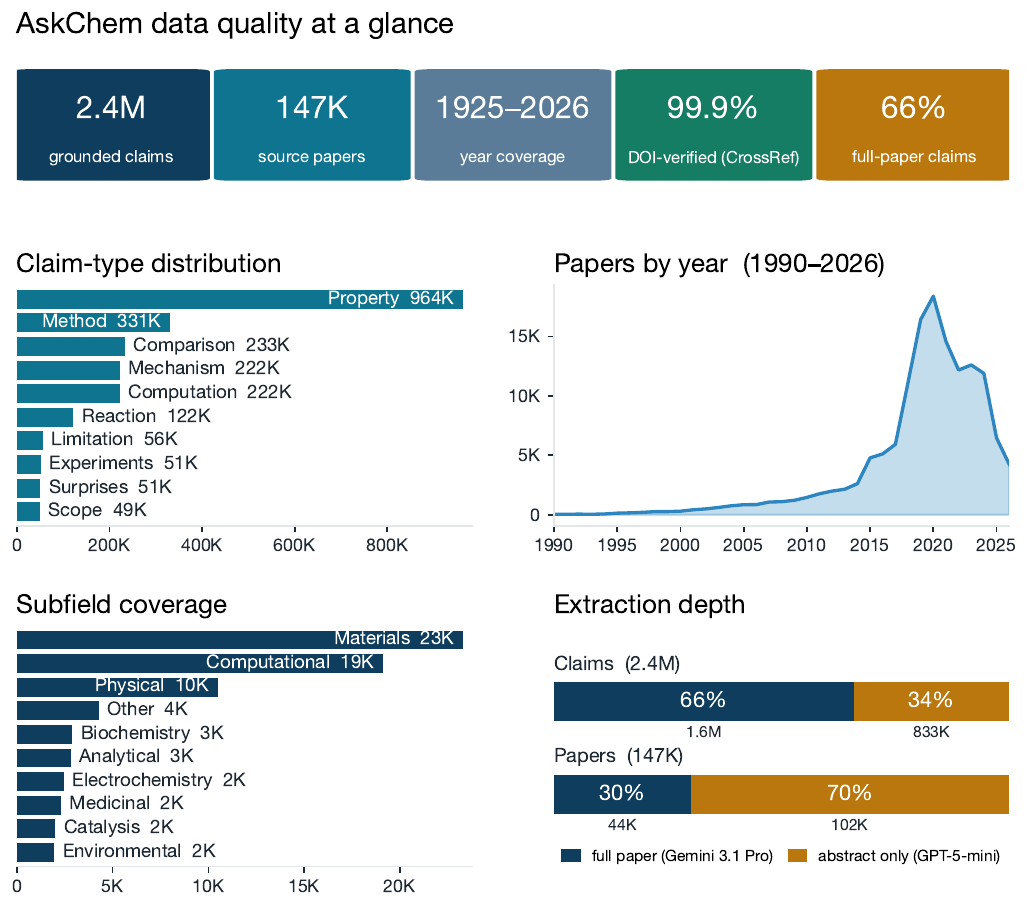}
  \caption{Corpus-scale coverage and automatic quality checks for the deployed index. These statistics characterize provenance and extraction depth; they do not replace expert judgments of claim semantics or taxonomy placement.}
  \label{fig:pipeline-quality}
\end{figure*}

\begin{figure*}[t]
  \centering
  \includegraphics[width=\linewidth]{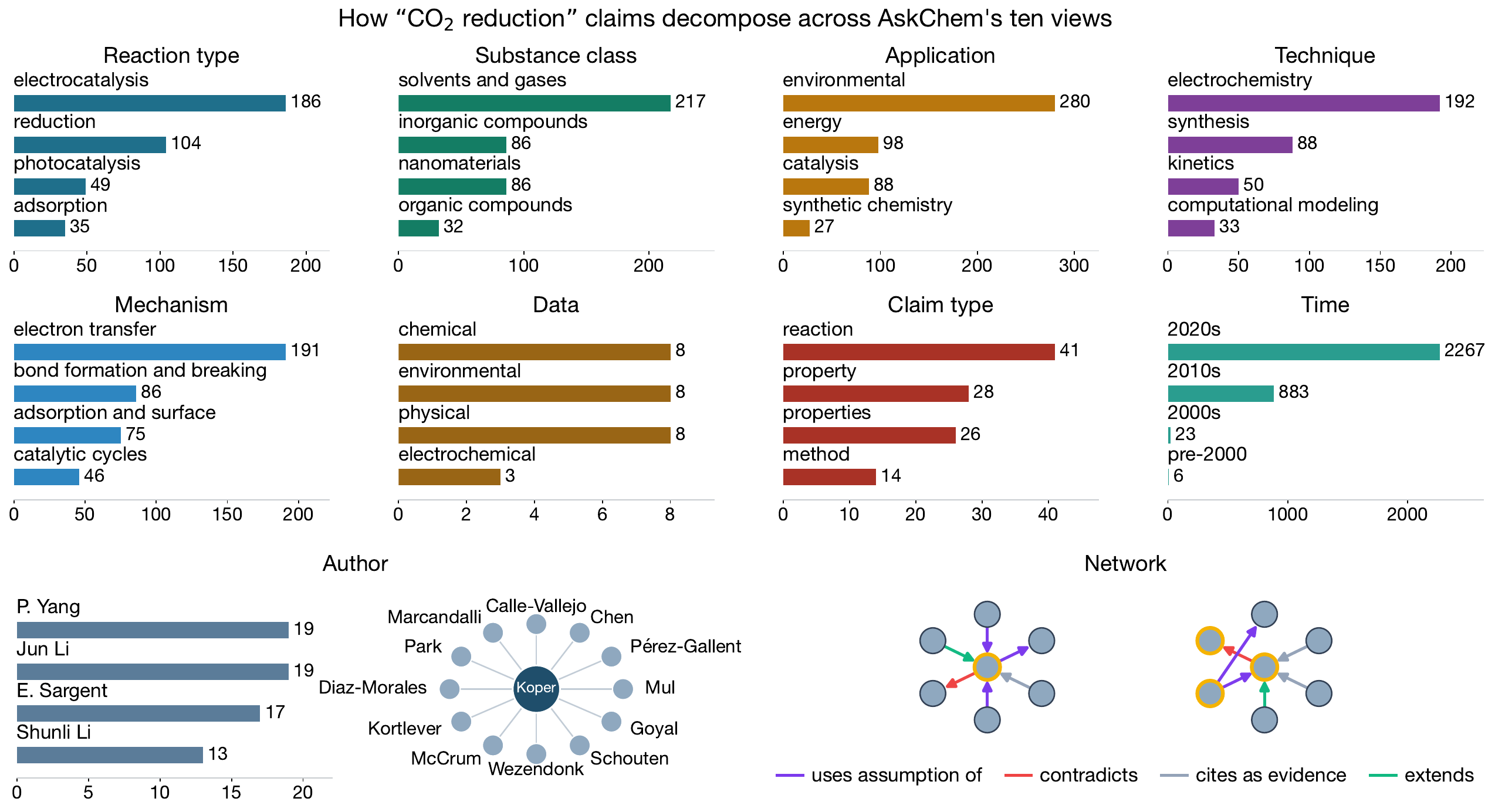}
  \caption{The same ``CO$_2$ reduction'' claims across 10 views: reaction, substance, application, technique, mechanism, data, claim type, and time panels expose faceted structure; author view has a coauthor relationship graph that exposes collaborations; network view displays the evidence graph.}
  \label{fig:topicviews}
\end{figure*}

\section{Claim-Centered Representation}
\label{sec:representation}

Figure~\ref{fig:architecture} shows the AskChem platform. The central object is a \textbf{Claim}: an atomic, typed scientific assertion extracted from a paper and grounded by a source DOI and a verbatim quote. A claim also includes structured fields, such as reactants, conditions, measurements, or materials, together with an extraction confidence score. This representation makes the individual finding, rather than the full paper, the object that can be searched, grouped, linked, and verified.

AskChem stores several structures under the same claim identity. A \textbf{Source} records paper-level metadata, including DOI, venue, year, citation count, and authors disambiguated via OpenAlex~\citep{openalex}. A \textbf{TreeNode} places a claim in one or more faceted taxonomy paths, while an \textbf{Edge} records a typed relation between two claims. Keeping these structures anchored to shared claim identifiers lets search, hierarchical browsing, and graph traversal return the same provenance-bearing objects rather than disconnected summaries.

The current live index contains \textbf{2.4M claims} from \textbf{147K papers} spanning 1925--2026, with \textbf{307K} populated taxonomy nodes (Figure~\ref{fig:pipeline-quality}). The data are stored in SQLite with FTS5 full-text search and a vector index, and are served through a FastAPI application used by the web interface and agent-facing APIs. Figure~\ref{fig:website} and Appendix~\ref{sec:appendix-construction} show example claims in the interface and index record.

\section{Claim Extraction and Evidence Graph}
\label{sec:extraction}

\paragraph{Extracting provenance-carrying claims.}
AskChem populates the claim store with two complementary extraction pipelines (Figure~\ref{fig:pipeline-quality}). A high-throughput extractor processes abstracts at scale, while a deeper extractor reads full-text PDFs, capturing claim types that are often absent from abstracts, such as hypotheses, limitations, and surprising findings. Each extraction call returns structured JSON that is validated against the claim schema, including required provenance fields, numeric ranges, and chemistry-specific fields.

These checks establish traceability rather than full semantic correctness. In the index, \textbf{100\%} of the 2.4M claims are source-grounded, carrying a claim type, a source DOI, and a verbatim quote.

\paragraph{Linking claims with an evidence graph.}
Cross-paper search often requires more than retrieving isolated findings: users also need to know how findings support, extend, derive from, or contradict one another. AskChem therefore adds a relation-extraction layer over the claim store. The extractor emits typed, directed edges between claims, including \texttt{cites\_as\_evidence}, \texttt{supports}, \texttt{extends}, \texttt{contradicts}, and \texttt{derives\_from}, with confidence scores and provenance.

The current graph contains \textbf{171{,}342} typed edges. A domain-expert author manually verified a stratified sample of 148 edges. After excluding 2 undecidable cases, 143 of 146 decidable edges had the correct relation type, giving an edge-type precision of \textbf{97.9\%}. The graph is used as a relational layer over retrieval rather than as a replacement for search: \texttt{/api/claims/\{id\}/neighborhood} returns inbound and outbound evidence links for a claim, and the web interface induces a graph over top search hits so users can move from a finding to claims that support, extend, or contradict it.

\begin{figure*}[t]
  \centering
  \includegraphics[width=0.65\linewidth]{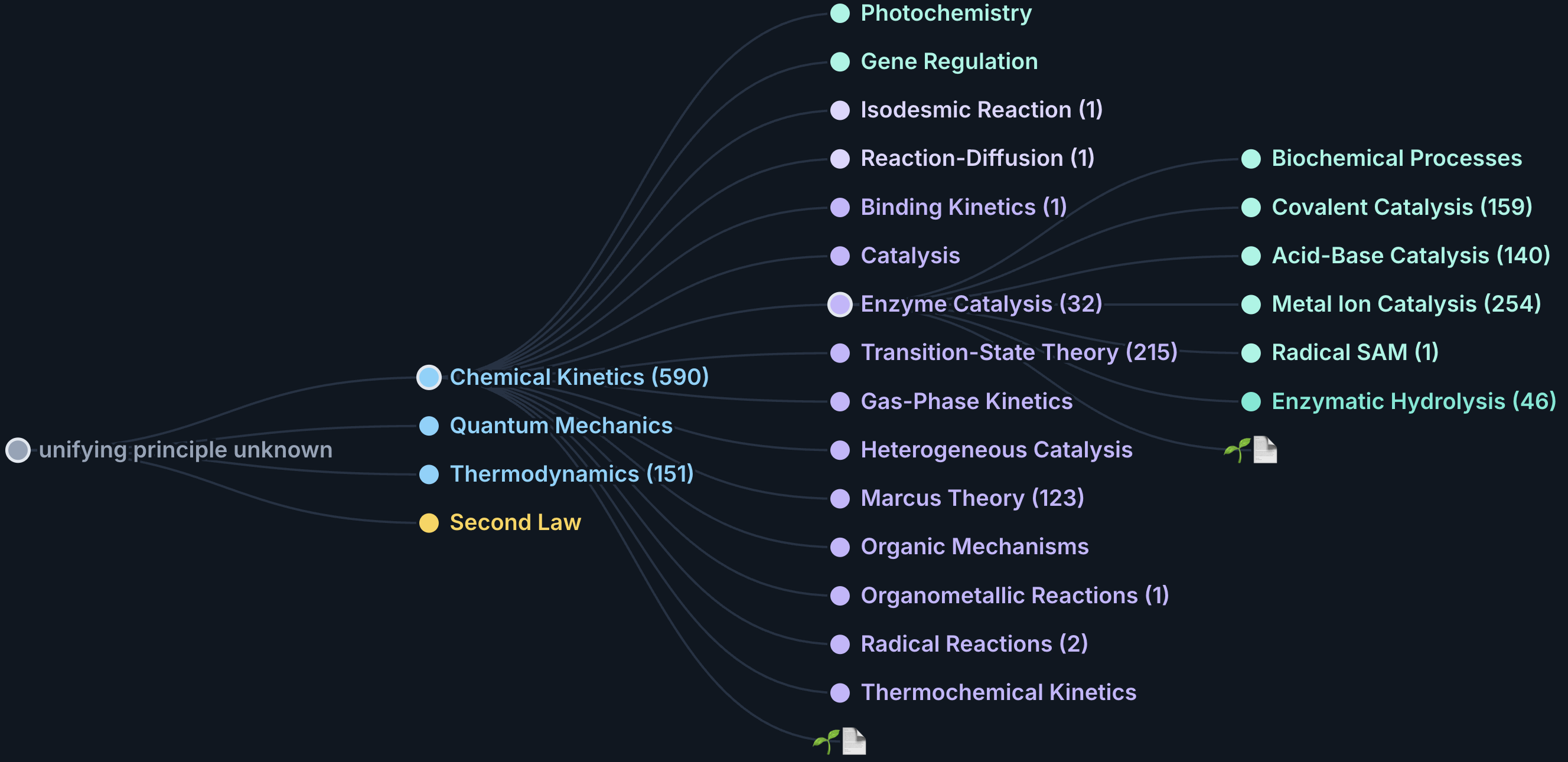}
  \caption{The principle-centered living taxonomy answers what governs a contribution. Unlike the stabilized faceted taxonomy used for search, it organizes indexed papers under principles, theories, models, and mechanisms.}
  \label{fig:livingtree}
\end{figure*}

\section{Stabilized Faceted Taxonomy}
\label{sec:organize}

\paragraph{Inducing stable search facets.}
AskChem uses a faceted taxonomy to organize claims by what they concern. Rather than imposing a fully predefined chemistry ontology, AskChem induces category paths while digesting papers and extracting claims. These paths reveal recurring terminology across reactions, substances, applications, techniques, and mechanisms. We then stabilize the induced paths through canonical top-level routing, synonym normalization, and fuzzy clustering of near-duplicate subcategories. The result is a corpus-derived but persistent set of L1/L2/L3 paths suitable for production retrieval and browsing.

\paragraph{Operational views over claims.}
The taxonomy defines multiple operational views over the same claim store. Each claim can be assigned a 2--5 segment path in populated views, such as \texttt{coupling/cross\_coupling/suzuki} under \emph{by\_reaction\_type}. Five content views capture what a claim is about: reaction type, substance class, application, technique, and mechanism topic. Additional views organize claim type, extracted measurements (data), time, and source authors. These facets are not treated as independent datasets; they are different navigational lenses over the same provenance-carrying claims. Figure~\ref{fig:topicviews} illustrates how one topic decomposes differently across these views. The last view is the evidence graph (Network) discussed in Section~\ref{sec:extraction}.

\paragraph{Retrieval and browsing.}
The stabilized taxonomy is used as an operational index, not only as a visualization. Hybrid \texttt{/search} combines FTS5 claim-text retrieval, paper-level recall, taxonomy-node recall, and dense-vector recall using reciprocal rank fusion~\citep{cormack2009rrf}. Returned claims retain their associated view paths, allowing UI and API clients to group results, expand related categories, or move from a retrieved claim into its hierarchy. Browse endpoints expose the same taxonomy nodes with claim counts and temporal overlays. In this design, lexical and semantic retrieval find candidate evidence, while faceted structure broadens, groups, and contextualizes that evidence under stable chemistry concepts.

\section{Exploratory Living Taxonomy}
\label{sec:living}


\paragraph{Principle-centered organization.}
The stabilized faceted taxonomy asks what a claim is about; the \textbf{living taxonomy} asks which broader scientific idea governs a paper's contribution. It organizes paper-grounded leaves under principles, theories, models, mechanisms, and phenomena (Figure~\ref{fig:livingtree}). The current tree contains 4{,}931 nodes and covers 1.1M claims across 361K paper placements, with an abstention mechanism that proposes new branches when no existing node is suitable. We treat it as an exploratory overview of the indexed corpus rather than a fully validated scientific ontology. The construction details appear in Appendix~\ref{sec:appendix-construction}.

\section{Cross-Paper Search and Demonstration}
\label{sec:human}

\paragraph{One query across three structures.} A researcher asks \emph{``what electrocatalysts reduce CO$_2$ to CO, and at what Faradaic efficiency?''} Hybrid search returns claim-level results with structured conditions, verbatim evidence, taxonomy paths, and source DOIs. The user can open a path to browse its hierarchy, group hits into an evidence graph, or situate a paper in the living taxonomy. Human and agent interfaces retrieve the same claim identities through the web UI, REST, SDK, or MCP \citep{mcp}; secondary workflows and endpoint examples appear in Appendix~\ref{sec:appendix-construction}.

\section{Evaluation}
\label{sec:eval}


We evaluate AskChem along four questions aligned with the system design:
whether extracted claims remain traceable to source evidence (\textbf{RQ1});
whether claim-level structure is reliable enough to support navigation (\textbf{RQ2});
whether claim-centered retrieval improves cross-paper chemistry answers (\textbf{RQ3});
and whether the deployed system operates at corpus scale (\textbf{RQ4}).


\paragraph{RQ1: Are claims source-grounded?}
Figure~\ref{fig:pipeline-quality} summarizes provenance checks for the deployed corpus.
In the current index, 100\% of claims are source-grounded (claim type, source DOI, and a verbatim quote).
These checks establish traceability and help detect unsupported generated text, but they do not prove that every extracted claim semantically interprets its source correctly.


\paragraph{RQ2: Are claim-level structures reliable and useful?}
The evidence graph is evaluated by the domain-expert audit reported in Section~\ref{sec:extraction}, which estimates 97.9\% edge-type precision.
The stabilized faceted taxonomy is used operationally in production: it contributes one of the retrieval signals in hybrid search and drives browsing, grouping, and temporal overlays in the UI.
However, we do not yet isolate the retrieval gain from taxonomy-based recall or report expert validation of taxonomy placement.
The living taxonomy is likewise treated as exploratory rather than as a fully validated scientific ontology.

\begin{table*}[t]
  \centering
  \small
  \begin{tabular}{lccccc}
    \toprule
    \textbf{Metric} & \textbf{LLM only} & \textbf{+AskChem} & \textbf{+Paperclip} & \textbf{Edison Scientific} & \textbf{NotebookLM} \\
    \midrule
    DOI existence (\%)        & 88.3 & \textbf{100} & \textbf{100} & 99.1 & 93.7 \\
    Citation density (/ans.)  & 9.6  & \textbf{18.1} & 7.5 & 10.7 & 7.9 \\
    Grounded specificity      & 8.1  & 5.9 & 0.5 & \textbf{29.2} & 0.1 \\
    Recent high-impact (\%)   & 0.6  & \textbf{18.5} & 6.1 & 11.3 & 12.1 \\
    Paper relevance (0--3)    & 1.66 & \textbf{2.15} & 1.72 & 2.07 & 1.84 \\
    On-topic $\geq 2$ (\%)    & 65.8 & 86.6 & 57.8 & \textbf{89.7} & 78.9 \\
    \bottomrule
  \end{tabular}
  \caption{AskChem-Bench: 30 cross-paper chemistry questions (condition aggregation, temporal tracking, contradiction surfacing), GPT-5.5 reader; all five systems cover the full 30. DOIs are verified via CrossRef. Bold marks the best value in each row; Paperclip ties AskChem only on DOI existence.}
  \label{tab:bench}
\end{table*}

\begin{figure*}[t]
  \centering
  \small
  \fbox{\begin{minipage}[t]{0.45\textwidth}
    \textbf{GPT-5.5 alone} --- fabricates citations\\[2pt]
    \itshape ``Au nanoneedles: $\sim$95\% FE for CO, $\sim$22\,mA\,cm$^{-2}$ CO partial current\ldots''\normalfont~[\texttt{\sout{10.1021/jacs.7b00392}}]\\[3pt]
    \textbf{6 of 14} cited DOIs do not resolve on CrossRef; the specific values cannot be verified.
  \end{minipage}}\hfill
  \fbox{\begin{minipage}[t]{0.45\textwidth}
    \textbf{GPT-5.5 + AskChem} --- grounded in retrieved claims\\[2pt]
    \itshape ``Atomically dispersed Fe$^{3+}$ sites: overpotential 360\,mV, $j_{\mathrm{CO}}=20$\,mA\,cm$^{-2}$''\normalfont~(\texttt{10.1126/science.aaw7515})\\[3pt]
    \textbf{All 22} cited DOIs verify; reported values are verbatim from source abstracts.
  \end{minipage}}
  \caption{Same question (AskChem-Bench ca04, \emph{``electrocatalysts for CO$_2$ reduction to CO/formate''}), answering alone, GPT-5.5 fabricates 6 of 14 DOIs; when grounded in AskChem's retrieved claims, all 22 citations resolve.}
  \label{fig:bench-example}
\end{figure*}



\paragraph{RQ3: Does claim-centered retrieval improve cross-paper synthesis?}
AskChem-Bench contains 30 cross-paper chemistry questions spanning condition aggregation, temporal tracking, and contradiction surfacing.
We compare five settings: a GPT-5.5 reader without retrieval, the same reader grounded through AskChem, Paperclip~\citep{paperclip2026}, Edison Scientific's PaperQA-family agent~\citep{paperqa2}, and Google NotebookLM Deep Research~\citep{notebooklm2025}.
Appendix~\ref{sec:appendix-bench} presents the full questions, retrieval budgets, metrics, DOI verification procedure, and judge calibration.

Table~\ref{tab:bench} reports the results.
Grounding the reader in AskChem yields 100\% resolvable DOIs, compared with 88.3\% without retrieval, and achieves the highest citation density among the five settings at 18.1 verified DOIs per answer.
AskChem also obtains the best mean relevance score and the highest coverage of recent high-impact work.
Figure~\ref{fig:bench-example} shows a representative case in which the reader alone fabricates citations, while the AskChem-grounded reader cites only resolvable DOIs.

The comparison also highlights AskChem's limitations.
Edison Scientific, a closed agentic deep-research system, produces substantially more citation-linked quantitative detail and a slightly higher on-topic rate.
AskChem has a different profile: it is claim-level, open-data, fast enough for interactive browsing, and directly usable as an agent tool, while still eliminating DOI hallucination on this benchmark.

\paragraph{RQ4: Does AskChem operate at corpus scale?}
The deployed service jointly queries 2.4M claims, 307K taxonomy nodes, and 171K evidence edges through the same REST schema used by the web interface, SDK, and MCP server.

\section{Comparison with Existing Systems}
\label{sec:related}

\paragraph{Databases and search.} Reaxys \citep{reaxys} and SciFinder \citep{scifinder} provide proprietary curated reaction/substance data; PubChem \citep{pubchem} and ChEMBL \citep{chembl} openly structure molecules and bioactivities. Semantic Scholar \citep{semantic_scholar} and Google Scholar \citep{google_scholar} contain rich metadata but primarily return ranked papers. AskChem instead exposes faceted structure over provenance-carrying narrative claims.

\paragraph{Scientific NLP and assistants.} Claim extraction and graph construction build on scientific information and relation extraction \citep{luan2018multi}; the hierarchies relate to taxonomy induction \citep{taxonomy_induction,zeng2024chainoflayer}, and retrieval combines RAG, sentence embeddings, and rank fusion \citep{lewis2020rag,reimers2019sbert,cormack2009rrf}. Unlike the answer-generating assistants benchmarked in Section~\ref{sec:eval}, AskChem exposes a persistent claim store as reusable human- and agent-facing infrastructure.

\section{Conclusion}


AskChem demonstrates a claim-centered alternative to document-only chemistry literature search. By making provenance-carrying claims the central unit, AskChem supports search results that can be verified against source quotes and DOIs, organized through stabilized facets, and connected through evidence relations across papers. The deployed system operates at corpus scale and exposes the same claim objects to both human users and AI agents. On cross-paper chemistry questions, AskChem improves citation groundedness, suggesting that claim-centered infrastructure is a practical foundation for chemistry literature synthesis.

\section{License and Availability}
\label{sec:avail}

AskChem is live at \url{https://askchem.org} and a screencast demo is at \url{https://youtu.be/SOjueOlPS-8}. The MIT-licensed source code is at \url{https://github.com/bingyan4science/askchem}, and the CC-BY index snapshot is released at \url{https://huggingface.co/datasets/bing-yan/askchem}. OpenAPI documentation, benchmark artifacts, and SDK/MCP access are listed in Appendix~\ref{sec:appendix-construction}.

\section*{Limitations}

The corpus covers only a fraction of chemistry, and abstract extraction is shallower than full-text extraction. LLM-generated claims, relations, and taxonomy placements can be wrong; AskChem-Bench measures groundedness on 30 questions rather than full factual accuracy or user utility.
String-based taxonomy normalization can merge distinct categories or retain near-duplicates, and its retrieval gain has not been isolated.

\section*{Ethics and Broader Impact}


AskChem uses LLM-based extraction, and its claims may contain errors. Each claim therefore includes provenance: a source DOI and a verbatim quote, so users can verify it against the original paper. The interface also supports community flagging. AskChem is intended to assist literature search and synthesis, not to replace reading primary sources for critical decisions. By grounding AI-assisted chemistry workflows in verifiable citations, it aims to reduce citation fabrication and make generated answers easier to audit. The system indexes papers using public metadata, abstracts, and open-access full text where available. The public index exposes claim text, source metadata, and provenance, but does not redistribute paywalled full text.

\section*{Acknowledgements}
We thank NYU HPC for their generous support and help in computational simulations and experiments. We are grateful to the NYU Google Cloud Platform (GCP) for Research Grant Program, which provided the primary computational support for LLM-based extraction and processing in AskChem. This work was supported by the Institute of Information \& Communications Technology Planning \& Evaluation (IITP) with a grant funded by the Ministry of Science and ICT (MSIT) of the Republic of Korea in connection with the Global AI Frontier Lab International Collaborative Research. This work was partly supported in part by the NYUAD Center for Interdisciplinary Data Science \& AI (CIDSAI), funded by Tamkeen under the NYUAD Research Institute Award CG016.

\bibliography{references}

\clearpage
\appendix

\begin{table*}[t]
\centering
\small
\begin{tabular}{lll}
\toprule
\textbf{CA: conditions} & \textbf{TC: evolution} & \textbf{CS: conflicts} \\
\midrule
C--N coupling & Perovskite degradation & Ag nanoparticle safety \\
Suzuki--Miyaura & MOF aqueous stability & DFT functional accuracy \\
Asym.\ hydrogenation & Li-ion SEI formation & MOF practical stability \\
CO$_2$ reduction & Protein corona & Solvent effects \\
Water splitting & TiO$_2$ photocatalysis & Interface passivation \\
ROMP & CO$_2$RR mechanisms & PLA biodegradability \\
Alcohol oxidation & Drug delivery & N-doped graphene \\
Heck coupling & Graphene defects & Cu for CO$_2$RR \\
Biocatalysis & Nanoparticle toxicity & Solid electrolyte safety \\
Metal adsorption & Electrolyte stability & Nanoparticle size effects \\
\bottomrule
\end{tabular}
\caption{AskChem-Bench topics (30 questions total).}
\label{tab:bench-topics}
\end{table*}

\section{Benchmark and Evaluation Details}
\label{sec:appendix-bench}
\paragraph{Question bank.} AskChem-Bench v1.1 contains ten questions in each of three cross-paper tasks. Table~\ref{tab:bench-topics} abbreviates the topics; the public benchmark endpoint releases their full wording.

\paragraph{Protocol.} GPT-5.5 answers all 30 questions in five settings. AskChem rewrites each question into 3--4 keyword subqueries, fans them out to hybrid search, and diversifies the merged evidence to at most 40 claims before grounded synthesis. Paperclip uses the same rewriter and synthesizer over its paper retrieval; Edison Scientific uses its PaperQA-family agent, and NotebookLM uses Deep Research. AskChem is judged against the claims it exposes, whereas paper-level systems are judged against cited titles and abstracts. Every extracted DOI is checked through CrossRef. The Gemini 3.1 Pro relevance judge was calibrated on 100 domain-expert labels (93\% agreement; $\kappa=0.914$).

\paragraph{Metrics.}

\begin{center}
\small
\setlength{\tabcolsep}{3pt}
\begin{tabular}{@{}>{\raggedright\arraybackslash}p{0.29\columnwidth}>{\raggedright\arraybackslash}p{0.62\columnwidth}@{}}
\toprule
\textbf{Metric} & \textbf{Definition} \\
\midrule
DOI existence & Fraction resolving in CrossRef. \\
Citation density & Distinct verified DOIs per answer. \\
Grounded specificity & Quantitative tokens sharing a sentence with a citation marker. \\
Recent impact & Cited papers from the last five years with $\geq 50$ citations. \\
Relevance & Judge score: 3 direct, 2 on-topic, 1 loose, 0 irrelevant. \\
\bottomrule
\end{tabular}
\end{center}

\paragraph{Reproducibility.} \url{https://askchem.org/api/benchmark} provides the full questions, methodology, aggregate and task-level results, and recorded index snapshots. Prompt templates and the rerun script accompany the source release; the indexed data is at \url{https://huggingface.co/datasets/bing-yan/askchem}.

\section{Representation and Structural Construction}
\label{sec:appendix-construction}

\paragraph{Claim record.} A real claim from the index (\texttt{claim\_id 7c92fcacd8cb64d4}), abbreviated only for width:
\begin{quote}\footnotesize\ttfamily
\{"claim\_type":"reaction",\\
\hspace*{0.7em}"source\_doi":"10.1002/anie.201914977",\\
\hspace*{0.7em}"reaction\_type":"electrocatalytic\\
\hspace*{2.6em}CO2 reduction (to CO)",\\
\hspace*{0.7em}"reactants":[\{"name":"CO2","role":\\
\hspace*{2.6em}"substrate"\},\{"name":"Ni SA-N2-C",\\
\hspace*{2.6em}"role":"catalyst"\}],\\
\hspace*{0.7em}"products":[\{"name":"CO"\}],\\
\hspace*{0.7em}"outcomes":\{"selectivity":\\
\hspace*{2.6em}"CO Faradaic efficiency 98\%",\\
\hspace*{2.6em}"other":"turnover frequency 1622 h-1"\},\\
\hspace*{0.7em}"verbatim\_quote":"the Ni SA-N2-C\\
\hspace*{2.6em}catalyst, with the lowest N\\
\hspace*{2.6em}coordination number, achieves very\\
\hspace*{2.6em}high CO Faradaic efficiency (98\%) and\\
\hspace*{2.6em}turnover frequency (1622 h-1),",\\
\hspace*{0.7em}"view\_paths":\{"by\_reaction\_type":\\
\hspace*{2.6em}["electrocatalysis","co2\_reduction"], \ldots\}\}
\end{quote}

Structured full-paper claims that lack a contiguous quote instead carry an \texttt{evidence\_locator} (\texttt{location\_in\_paper} plus structured evidence), so every claim is grounded by a quote or a locator.

\paragraph{Evidence edge.} A real typed edge from \texttt{claim\_edges} links claims across papers:
\begin{quote}\footnotesize\ttfamily
\{"edge\_type":"supports",\\
\hspace*{0.7em}"to\_doi":"10.1021/acs.joc.9b01692",\\
\hspace*{0.7em}"confidence":"high",\\
\hspace*{0.7em}"evidence":"The observed selective coupling\\
\hspace*{2.6em}at the chloride position using SIPr\\
\hspace*{2.6em}directly provides evidence for the\\
\hspace*{2.6em}overarching claim of high\\
\hspace*{2.6em}ligand-controlled selectivity."\}
\end{quote}

\paragraph{Extractor details.}
The abstract extractor uses \textbf{GPT-5-mini} over paper titles and abstracts; the deep full-text extractor uses \textbf{Gemini 3.1 Pro} with native-PDF input via Vertex AI batch. A small legacy slice of the index predates this pipeline (GPT-4o / GPT-4o-mini). All calls use JSON-object-constrained decoding (\texttt{response\_format=\{"type":"json\_object"\}}) at the provider-default temperature, with automatic retry on invalid JSON or schema-invalid output. Prompts and schemas are released with the source code.

\paragraph{Stabilized faceted taxonomy.} Category paths were induced while digesting papers and extracted claims, then consolidated through canonical L1 routing, synonym normalization, and fuzzy clustering into persistent L1/L2/L3 categories. Five content views cover reaction, substance, application, technique, and mechanism topic; claim-type, measurement, and time views add complementary facets. A path such as \texttt{coupling/cross\_coupling/suzuki} is both a browse location and a taxonomy-recall signal in hybrid search.

\paragraph{Evidence graph.} A second pass emits directed \texttt{supports}, \texttt{contradicts}, \texttt{extends}, \texttt{derives\_from}, and \texttt{cites\_as\_evidence} edges with confidence and provenance. The expert audit sampled 148 edges across relation types: 146 were decidable and 143 had the correct type (97.9\% precision); two were excluded as undecidable.

\paragraph{Living Taxonomy.} This exploratory hierarchy is intentionally explanatory rather than faceted. A 4{,}931-node tree of principles, theories, models, mechanisms, and phenomena anchors paper-grounded leaves, of which 663 are open \texttt{proposed} branches. An LLM reads each paper's claims and names the host that best governs its contribution; if none fits, it abstains through a \texttt{proposed} branch.

\begin{table}[h]
  \centering
  \footnotesize
  \setlength{\tabcolsep}{3pt}
  \begin{tabular}{@{}lrr@{}}
    \toprule
    Living Taxonomy view & Paper placements & Claims \\
    \midrule
    Substance class & 111{,}714 & 490{,}137 \\
    Technique & 119{,}361 & 292{,}154 \\
    Mechanism & 98{,}665 & 217{,}388 \\
    Reaction type & 30{,}806 & 62{,}294 \\
    \midrule
    \textbf{Total} & \textbf{360{,}546} & \textbf{1{,}061{,}973} \\
    \bottomrule
  \end{tabular}
  \caption{Coverage of the exploratory Living Taxonomy. A paper may appear in multiple views, so placements are not unique-paper counts.}
  \label{tab:living}
\end{table}

\paragraph{Placement evidence.} Nitrite reduction is placed under electrocatalytic redox and diene cross-metathesis under olefin metathesis, whereas nearest-neighbor placement can force-fit low-margin cases. These examples motivate the design but are not a full evaluation; expert placement validation remains future work.

\paragraph{Validation gates.} Extraction responses must parse against the claim schema and contain required provenance fields; numeric and chemical fields are checked when present. Faceted paths are routed to canonical top-level categories before lower-level normalization. Evidence edges retain extractor confidence and provenance, while Living Taxonomy placements can abstain rather than force an unsupported host.


\paragraph{End-to-end API path.} The same objects used by the web demo are available over REST:
\begin{quote}\footnotesize\ttfamily
GET /api/search?q=CO2+reduction\\
GET /api/claims/\{claim\_id\}\\
GET /api/claims/\{claim\_id\}/neighborhood\\
GET /api/sources/\{doi\}
\end{quote}
Search returns claim IDs and taxonomy paths; claim and neighborhood lookup expose provenance and evidence edges; source lookup returns all indexed claims from the paper. The same API supports reading lists, author networks, saved searches, and subscriptions; SDK and MCP wrappers expose it to agents. Complete schemas are at \url{https://askchem.org/api/docs}, benchmark artifacts at \url{https://askchem.org/api/benchmark}, and the screencast demo is at \url{https://youtu.be/SOjueOlPS-8}.

\end{document}